\documentclass{article}

\usepackage[preprint,nonatbib]{neurips_2020}
\usepackage[
    sorting=none, style=numeric-comp,
    ]{biblatex}
\usepackage{tabularx}
\usepackage[utf8]{inputenc} 
\usepackage[T1]{fontenc}    
\usepackage{hyperref}       
\usepackage{url}            
\usepackage{booktabs}       
\usepackage{amsfonts}       
\usepackage{nicefrac}       
\usepackage{microtype}      
\usepackage{graphicx}
\usepackage{amsmath}

\title{$\xi$-torch: differentiable scientific computing library}

\author{%
  Muhammad F. Kasim and Sam M. Vinko \\
  Department of Physics, University of Oxford\\
  Oxford, OX1 3PU \\
  \texttt{\{firstname.lastname\}@physics.ox.ac.uk} \\
}

\begin{document}

\maketitle

\begin{abstract}
  Physics-informed learning has shown to have a better generalization than learning without physical priors.
  However, training physics-informed deep neural networks requires some aspect of physical simulations to be written in a differentiable manner.
  Unfortunately, some operations and functionals commonly used in physical simulations are scattered, hard to integrate, and lack higher order derivatives which are needed in physical simulations.
  In this work, we present $\xi$-torch, a library of differentiable functionals for scientific simulations.
  Example functionals are a root finder and an initial value problem solver, among others.
  The gradient of functionals in $\xi$-torch are written based on their analytical expression to improve numerical stability and reduce memory requirements.
  $\xi$-torch also provides second and higher order derivatives of the functionals which are rarely available in existing packages.
  We show two applications of this library in optimizing parameters in physics simulations.
  The library and all test cases in this work can be found at \href{https://github.com/xitorch/xitorch/}{https://github.com/xitorch/xitorch/} and the documentation at \href{https://xitorch.readthedocs.io}{https://xitorch.readthedocs.io}.
\end{abstract}

\section{Introduction}

In the last decade, machine learning and deep learning has shown tremendous applications in scientific computing and simulations.
Deep learning has been employed to construct fast surrogate models of expensive simulations in various fields~\cite{kasim2020up}, to increase prediction accuracy by replacing inaccurate simulations with trainable neural networks~\cite{humbird2019transfer}, and also to discover new types of materials and drugs~\cite{rifaioglu2019recentdrugdiscovery}.
Despite the wide range of applications, the capability of deep learning in scientific simulations has not been fully unveiled.
One of the reasons is the requirement for large amounts of training data which might not be available for some applications.

A promising way to address the problem of large data requirement is by incorporating prior physics knowledge into deep learning.
Combining physics and deep learning can be done by using deep neural networks for expensive or inaccurate part of the simulations and writing the other parts in a differentiable manner for an efficient gradient propagation~\cite{ingraham2019learning}.
Using the combination of physics and deep neural networks removes the necessity of learning the known and accurate behaviours in the simulations, thus resulting in a better accuracy and generalization even with less data~\cite{de2018end}.

One major obstacles in combining scientific simulation with deep learning is the lack of libraries containing differentiable functionals commonly used in scientific simulations, such as an initial value problem solver (e.g. used intensively in molecular dynamics), self-consistency finder (e.g. in quantum chemistry~\cite{kohn1965self} and laser-matter interactions~\cite{vinko2020time}), and Monte Carlo calculations (e.g. in particle-matter interactions~\cite{collaboration2003geant4}).
Although some implementations of differentiable functionals are available~\cite{chen2018neuralode,bai2019deepeq}, they are typically scattered across different packages and automatic differentiation (AD) libraries, making it harder to integrate them into a single simulation package.
Moreover, most of the available packages provide only the first order derivatives and not higher order derivatives.
While higher order derivatives are not critical for pure machine learning applications, they play important roles in scientific simulations, e.g. calculating vibrational frequency requires second order derivatives.

Here we present a PyTorch-based~\cite{paszke2019pytorch} library that contains functionals commonly used for scientific simulations as well as their first and higher order derivatives.
The library is called  $\xi$-torch (read as \textit{scitorch}).
We drew a lot of inspirations from SciPy~\cite{virtanen2020scipy} and leverages the automatic differentiation (AD) advantage from PyTorch.
At the time of writing, $\xi$-torch is divided into several modules: \texttt{optimize} for optimization and root finding, \texttt{integrate} for quadratures and integrations, \texttt{interp} for interpolations, and \texttt{linalg} for linear algebra and sparse linear algebra operations.
Besides providing the functionals and their first order derivatives, $\xi$-torch also provides second and higher order derivatives of the implemented functionals which are rarely available in the existing libraries.
The source code of $\xi$-torch can be found at \href{https://github.com/xitorch/xitorch/}{https://github.com/xitorch/xitorch/}.

\section{Structure}

\subsection{Pure functions}

One of the main advantages of $\xi$-torch is the availability of second and higher order derivatives of the functionals.
An efficient way to make higher order derivatives available is by generating them using AD from the implementation of the first order derivative, which necessitates its correct implementation.
To write the correct first order implementation of functionals, we found that the functions in the functional inputs must be pure functions, i.e. functions where the outputs are fully determined by the inputs.

The requirement of having pure functions as the inputs of functionals raises difficulties in using them for stateful functions, such as an object's method.
To address this problem, $\xi$-torch provides an internal function to convert object's methods into pure functions by identifying which object's state variables are affecting the outputs and making them changeable.
This approach has been proven successful in providing the second order derivative of the implemented functionals, as checked by PyTorch's second order derivative checker function, i.e. \texttt{gradgradcheck}.

Efficiently identifying the state variables affecting a method's output is a challenging task.
For classes that are derived from \texttt{torch.nn.Module}, it can be done by using its \texttt{named\_parameters} method.
However, for a large class with many methods, $\xi$-torch provides \texttt{EditableModule} base class to be inherited.
Classes with \texttt{EditableModule} as a parent must implement a function \texttt{getparamnames} that manually lists variable names affecting the output of each method.
A debugging mode is provided by $\xi$-torch to check if the lists of variables are correct.

\subsection{Linear operator}

For some functionals, the first order derivative implementation requires operations with a Jacobian or Hessian matrix.
For example, the first derivative of root finder and minimizer, respectively, require the calculations of inverse Jacobian and Hessian matrices.
If the function takes a small number of inputs, the Jacobian and Hessian matrices can be constructed and the inverses can be calculated explicitly.
However, in some cases the number of inputs and outputs are large enough to render the matrix construction inefficient.

To avoid the inefficient matrix construction, $\xi$-torch provides a sub-library for sparse linear operator.
It provides a base class of \texttt{LinearOperator} that can be extended to any user-defined linear operators.
At the bare minimum, a custom \texttt{LinearOperator} only requires a matrix-vector multiplication defined.
Matrix-matrix multiplication is done by default using batched matrix-vector multiplications while the right-multiplication is performed using the adjoint trick with the help of PyTorch's AD engine (see appendix \ref{app:adjoint-trick} for details).
$\xi$-torch also provides a debugging mode which can check if the user-defined linear operator behaves correctly as a proper linear operator.

Besides the \texttt{LinearOperator} base class, $\xi$-torch provides several operations involving linear operators.
Currently those are: \texttt{solve} to solve a linear algebra equation and \texttt{symeig} to obtain $n$-lowest or uppermost eigenpairs of a real symmetric linear operator.
Those functions are designed to operate without the need to construct the matrix explicitly.
The details of those operations and their derivatives are presented in Appendix \ref{app:vjp}.

\subsection{\label{subsec:functional}Functionals and operations}

There are two distinct operations provided in $\xi$-torch: functionals and operations.
Functionals take other functions as inputs while operations do not.
Obtaining derivatives in operations (e.g. \texttt{Interp1D}) can be done just by writing the algorithms in a plain PyTorch, while calculating derivatives of functionals needs more effort and restrictions as described later.
The implemented functionals and operations at the time of writing are listed in Table \ref{tab:functional}.

Every function that goes into a $\xi$-torch functional will be converted internally into a pure function for the purpose of higher order derivatives.
Because of this, functionals in $\xi$-torch can only accept functions that are: (1) a pure function, (2) a method from a \texttt{torch.nn.Module} class, or (3) a method from an \texttt{EditableModule} class.

The API of the functionals are designed to be as intuitive as possible, taking the inspiration from SciPy.
As an example, finding the root of $f(x,a) = \sqrt{x} - a = 0$ can be done by the code below.
~
\hrule
\begin{verbatim}
func1 = lambda x,a: x**0.5 - a
a = torch.tensor(1., requires_grad=True)
x0 = torch.tensor(0.1, requires_grad=True)  # initial guess
xroot = rootfinder(func1, x0, params=(a,))  # func1(x0, *params)
\end{verbatim}
\hrule

For all functionals, the first order derivatives are implemented based on their analytical expressions instead of performing backpropagation on the implemented forward algorithms.
For example, the first order derivative of root finder functional is implemented by solving a linear equation involving the Jacobian of the input function, instead of backpropagating the Broyden algorithm implementation.
This gives more freedom in implementing the forward algorithms, allowing us to use existing libraries, such as SciPy.
Besides giving more freedom in the forward implementation, using the derivative's analytical expression significantly reduces the memory requirements, and increases the numerical stability.
The mathematical details of the derivatives of the functionals are presented in appendix \ref{app:vjp}.

\begin{table}[]
    \centering
    \begin{tabular}{||p{0.14\linewidth}|p{0.14\linewidth}|p{0.30\linewidth}|p{0.27\linewidth}||}
        \hline
        \textbf{Module} & \textbf{Functional} &
        \textbf{Common name} & \textbf{Mathematical expression} \\
        \hline\hline
        \texttt{linalg} & \texttt{solve} & Linear equation solver & $\mathbf{A\hat{X}-M\hat{X}E}=\mathbf{B}$ \\
        \hline
        \texttt{linalg} & \texttt{symeig} & Eigendecomposition for symmetric matrix &
        $\mathbf{A\hat{U}=M\hat{U}\hat{\Lambda}}$ \\
        \hline
        \texttt{linalg} & \texttt{svd} & Singular Value Decomposition &
        $\mathbf{A=\hat{U}\hat{S}\hat{V}}^H$ \\
        \hline
        \texttt{optimize} & \texttt{rootfinder} & Rootfinder & $\mathbf{f}(\mathbf{\hat{y}},\theta)=\mathbf{0}$\\
        \hline
        \texttt{optimize} & \texttt{equilibrium} & Equilibrium finder & $\mathbf{f}(\mathbf{\hat{y}}, \theta) = \mathbf{\hat{y}}$ \\
        \hline
        \texttt{optimize} & \texttt{minimize} & Function minimizer & $\mathbf{\hat{y}} = \arg\min_\mathbf{y}\mathbf{f}(\mathbf{y}, \theta)$ \\
        \hline
        \texttt{integrate} & \texttt{quad} & 1D quadrature & $\mathbf{\hat{y}} = \int_a^b \mathbf{f}(x,\theta)\ \mathrm{d}x$
        \\
        \hline
        \texttt{integrate} & \texttt{solve\_ivp} & Initial value problem solver & $\mathbf{\hat{y}} = \mathbf{y_0} + \int_{t_0}^{t_1} \mathbf{f}(t, \mathbf{y}, \theta)\ \mathrm{d}t$ \\
        \hline
        \texttt{integrate} & \texttt{mcquad} & Monte Carlo quadrature & $\mathbf{\hat{y}} = \frac{\int \mathbf{f}(\mathbf{x},\theta_f) p(\mathbf{x}, \theta_p)\ \mathrm{d}\mathbf{x}}{\int p(\mathbf{x},\theta_p)\ \mathrm{d}\mathbf{x}}$ \\
        \hline
        \texttt{integrate} & \texttt{SQuad} & Fixed-samples quadrature & \\
        \hline
        \texttt{interpolate} & \texttt{Interp1D} & Unsructured 1D interpolation & \\
        \hline
    \end{tabular}
    \caption{List of implemented functionals and operations in $\xi$-torch. Outputs are denoted by hats.}
    \label{tab:functional}
\end{table}

\section{Example cases}

\subsection{Mirror design}
\begin{figure}
    \centering
    \includegraphics[width=0.9\linewidth]{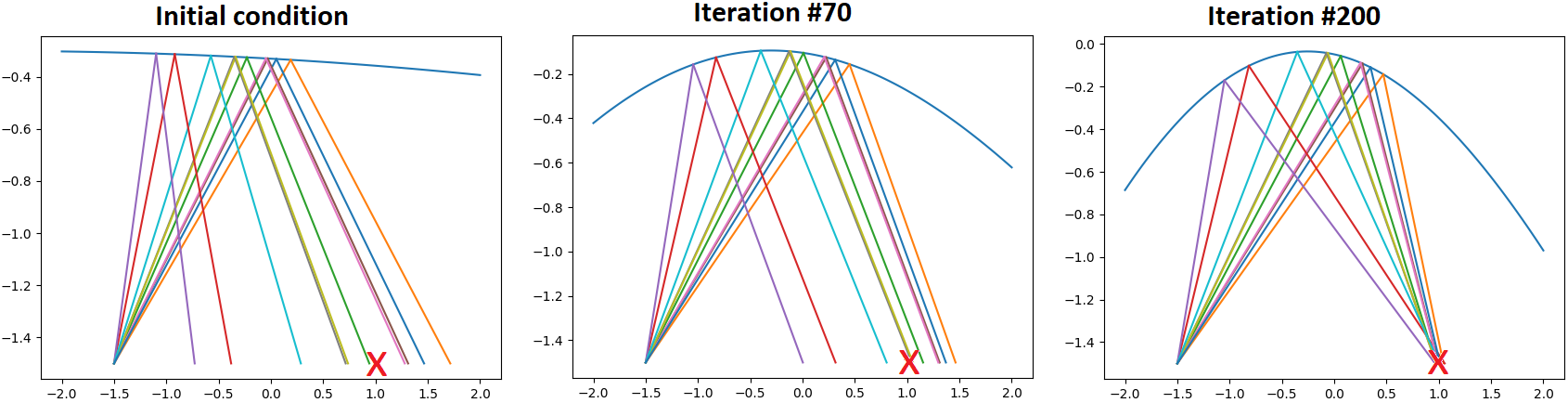}
    \caption{Illustration of the mirror design test case. The objective is to design a mirror shape to focus rays from a point source to a target point (marked by $\times$)}
    \label{fig:mirror-design}
\end{figure}

The first test case of $\xi$-torch is to design a mirror surface to focus rays of light.
Specifically, the mirror is located around $(0,0,0)$ with a point source at $(-1.5, 0, -1.5)$.
The objective of this test case is to focus the rays from the point source to the target point at $(1, 0, -1.5)$.
This test case requires the root finder functional to find the intersection between the rays and the mirror surface as well as its derivative in order to propagate the gradient to the mirror surface descriptor.
The illustration of the test case is shown in Figure \ref{fig:mirror-design}.

In this case, the mirror surface is described by a neural network consisting of 4 fully connected layers each followed by a softplus activation, except the last layer.
The neural network takes $x$ and $y$ as the inputs and produces $z$ as the output.
The loss function is calculated as the sum of square of the deviation of rays at $z=-1.5$ from the target position at $(1,0,-1.5)$.
The neural network parameters are varied to minimize the loss function using Adam optimizer~\cite{kingma2014adam} with a learning rate of $3\times10^{-4}$.

The results of the optimization can be seen in Figure \ref{fig:mirror-design}.
In 200 iterations, the optimizer already finds a reasonable mirror design to focus the rays at the target point.
This shows the capability of $\xi$-torch's root finder to propagate the gradient.

\subsection{Molecular dynamics}
\begin{figure}
    \centering
    \includegraphics[width=0.9\linewidth]{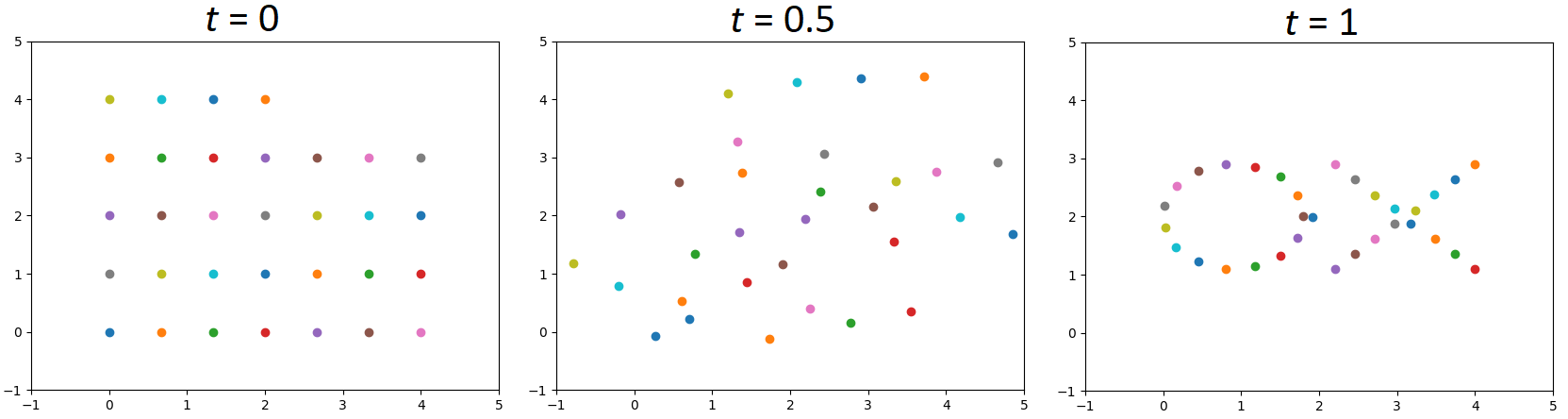}
    \caption{Snapshots of the optimized initial velocity in the molecular dynamics simulation}
    \label{fig:md}
\end{figure}

The second example is finding the initial velocities of an $n$-particle system so that they can form a certain shape at a certain time.
Two particles in the system with displacement $\mathbf{r}$ interact with force
\begin{equation}
    \mathbf{F}(\mathbf{r}) = -\left(|\mathbf{r}|^2 + a\right)^{-0.5}\hat{\mathbf{r}}.
\end{equation}
This force is similar to the gravitational force with an adjustment near $r=0$ to avoid singularity.
The adjustment value is $a = 0.1$.
This test case requires the \texttt{solve\_ivp} functional and its first order derivative.

In this case, 32 particles are arranged in a grid and given initial velocities at time $t=0$.
The initial velocities are optimized so that the particles are arranged as the letters "OX" at time $t=1$.
The optimization is performed using Adam algorithm~\cite{kingma2014adam} with learning rate $10^{-3}$.
The results after 7500 iterations can be seen in Figure \ref{fig:md}.

\section{Conclusion}

We have written a library of differentiable functionals for scientific computing and simulations based on PyTorch.
The derivatives of the functionals are implemented based on their analytical expressions which reduces memory requirements and increases numerical stability compared with performing backpropagation on the forward algorithm implementations.
The library also provides first and higher derivatives of the functionals which are commonly needed in scientific simulations.
With the availability of $\xi$-torch, we anticipate faster integration of deep learning and scientific simulations with benefits to both fields.


\section*{Broader impact statement}
$\xi$-torch contains differentiable operations and functionals for scientific computing that would help computational scientists to incorporate deep learning into scientific simulations.
Integrating deep learning in simulations can speed up and increase the accuracy of simulations by replacing expensive or inaccurate parts with trainable neural networks.
This could accelerate scientific discovery and help solving unsolved questions.
However, the approach of using deep learning to replace unfavorable parts of simulations could increase scientists' reliance on deep learning and discourage developments of new theories or computational algorithms.
As science is built on knowledge and understandings, excessive reliance on deep learning without the ability to interpret them would harm scientific developments in the long run.
Therefore, increasing applications of deep learning in science should be balanced by studies of interpreting deep neural network in science.

\small

\printbibliography

\clearpage

\appendix

\section{\label{app:notation}Appendix: partial derivative notations}
The partial derivative notation indicates a Jacobian matrix.
Therefore, for a vector $\mathbf{a}$ of size $n\times 1$, the notation $\partial \mathcal{L} / \partial \mathbf{a}$ would have a size of $1\times n$.
However, for a matrix, it will be written as $[\partial \mathcal{L} / \partial \mathbf{A}]$ which indicates it has the same size of $\mathbf{A}$.

\section{\label{app:adjoint-trick}Appendix: adjoint trick for right-multiplication}

In a linear operator where the matrix-vector left-multiplication (or simply ``multiplication'') is defined, we can calculate the right-multiplication automatically using automatic differentiation (AD).
For a linear operator $\mathbf{A}\in\mathbb{R}^{m\times n}$, the matrix-vector multiplication is written as
\begin{equation}
    \mathbf{y} = \mathbf{Ax}
\end{equation}
where $\mathbf{x}\in\mathbb{R}^{n\times 1}$ and $\mathbf{y}\in \mathbb{R}^{m\times 1}$ are vectors.

The Jacobian of the loss function with respect to $\mathbf{x}$ can be written as
\begin{equation}
    \frac{\partial \mathcal{L}}{\partial \mathbf{x}} = \frac{\partial \mathcal{L}}{\partial \mathbf{y}} \frac{\partial \mathbf{y}}{\partial \mathbf{x}}
\end{equation}
where $\partial\mathcal{L}/\partial\mathbf{x}\in \mathbb{R}^{1\times n}$, $\partial\mathcal{L}/\partial\mathbf{y}\in \mathbb{R}^{1\times m}$, and $\partial\mathbf{y}/\partial\mathbf{x} = \mathbf{A}$.
The calculation above can be done using the AD capability by supplying the value of $\partial\mathcal{L}/\partial\mathbf{y}$
Rearranging the previous equation makes
\begin{equation}
    \left(\frac{\partial \mathcal{L}}{\partial \mathbf{x}}\right)^T = A^T \left(\frac{\partial \mathcal{L}}{\partial \mathbf{y}}\right)^T.
\end{equation}
Therefore, using AD and a defined matrix-vector multiplication, the right-multiplication
\begin{equation}
    \mathbf{p} = \mathbf{A}^T \mathbf{q}
\end{equation}
can be computed by assigning $\mathbf{q} = (\partial\mathcal{L}/\partial\mathbf{y})^T$ in AD backward computation.

\section{\label{app:vjp} Appendix: vector-Jacobian products}

\subsection{\texttt{solve}}
The function \texttt{solve} solves the extended linear algebra equation to obtain an $n\times m$ vector of $\mathbf{X}$ from the following equation,
\begin{equation}
    \mathbf{AX-MXE} = \mathbf{B}
\end{equation}
where $\mathbf{A}$ and $\mathbf{M}$ are $n\times n$ linear operators with parameters $\theta_A$ and $\theta_M$, respectively, $\mathbf{B}$ is an $n\times m$ matrix, and $\mathbf{E}$ is an $m\times m$ diagonal matrix.
The equation above can be solved by employing root finder algorithms.

The vector-Jacobian product is given by
\begin{align}
    \left[\frac{\partial \mathcal{L}}{\partial \mathbf{B}}\right] : \mathrm{solve}\ \mathbf{A}^T\left[\frac{\partial \mathcal{L}}{\partial \mathbf{B}}\right] - \mathbf{M}^T \left[\frac{\partial \mathcal{L}}{\partial \mathbf{B}}\right] \mathbf{E} &= \left[\frac{\partial \mathcal{L}}{\partial \mathbf{X}}\right]\\
    \left[\frac{\partial\mathcal{L}}{\partial \mathbf{E}}\right] &= \mathrm{diag}\left(\left[\frac{\partial \mathcal{L}}{\partial \mathbf{B}}\right]^T\mathbf{MX}\right)\\
    \frac{\partial \mathcal{L}}{\partial \theta_A} &= -\mathrm{tr}\left[\left[\frac{\partial \mathcal{L}}{\partial \mathbf{B}}\right]^T \left(\frac{\partial \mathbf{A}}{\partial \theta_A}\right) \mathbf{X}\right]\\
    \frac{\partial \mathcal{L}}{\partial \theta_M} &= \mathrm{tr}\left[\left[\frac{\partial \mathcal{L}}{\partial \mathbf{B}}\right]^T \left(\frac{\partial \mathbf{M}}{\partial \theta_M}\right) \mathbf{XE}\right]\\
\end{align}
The first equation above can be solved using \texttt{solve} and transpose of $\mathbf{A}$ and $\mathbf{M}$.
Therefore, it requires the linear operators $\mathbf{A}$ and $\mathbf{M}$ to have transposed matrix-vector product defined.

\subsection{\texttt{symeig}}

Function \texttt{symeig} decomposes a linear operator to obtain $m$ lowest or uppermost eigenpairs,
\begin{equation}
    \mathbf{U},\mathbf{\Lambda}:\mathrm{where}\ \mathbf{AU} = \mathbf{MU\Lambda}
\end{equation}
where $\mathbf{A}$ and $\mathbf{M}$ are $n\times n$ real symmetric linear operators with parameters $\theta_A$ and $\theta_M$ respectively, $\mathbf{U}$ is an $n\times m$ eigenvectors matrix, and $\mathbf{\Lambda}$ is an $m\times m$ diagonal matrix containing the eigenvalues.

The vector-Jacobian products are given by
\begin{align}
    \mathbf{V}: \mathrm{solve}\ \mathbf{AV - MV\Lambda} &= \left[\frac{\partial \mathcal{L}}{\partial \mathbf{U}}\right] \\
    \mathbf{Y} &= \mathbf{V} - \mathbf{U}\mathrm{diag}(\mathbf{U}^T\mathbf{MV}) \\
    \frac{\partial \mathcal{L}}{\partial \theta_A} &= -\mathrm{tr} \left(\mathbf{Y}^T\frac{\partial \mathbf{A}}{\partial \theta_A}\mathbf{U}\right)\\
    \frac{\partial \mathcal{L}}{\partial \theta_M} &= \mathrm{tr}\left(\mathbf{Y}^T \frac{\partial \mathbf{M}}{\partial \theta_M} \mathbf{U}\mathbf{\Lambda}\right)
\end{align}

The equation to obtain $\mathbf{V}$ can be solved using the \texttt{solve} function.

\subsection{\label{app:rootfinder}\texttt{rootfinder}, \texttt{equilibrium}}

The functional \texttt{rootfinder} solves the rootfinding equation,
\begin{equation}
    \mathbf{y}: \mathrm{where}\ \mathbf{0} = \mathbf{f}(\mathbf{y}, \theta)
\end{equation}
where $\mathbf{y}$ is a vector of size $n\times 1$ (in practice, it can be a tensor of any size), $\theta$ represents other parameters, and $\mathbf{f}$ is a function which gives an output of the same size as $\mathbf{y}$.

The vector-Jacobian product is given by
\begin{align}
    \mathbf{h}^T &= \frac{\partial \mathcal{L}}{\partial \mathbf{y}} \left(\frac{\partial \mathbf{f}}{\partial \mathbf{y}}\right)^{-1} \\
    \frac{\partial \mathcal{L}}{\partial \theta} &= -\mathbf{h}^T \frac{\partial \mathbf{f}}{\partial \theta}
\end{align}
The first equation above can be calculated using \texttt{solve} using the transpose of Jacobian of $\mathbf{f}$ with respect to $\mathbf{y}$.

Functional \texttt{equilibrium} solves the equation
\begin{equation}
    \mathbf{y} = \mathbf{g}(\mathbf{y}, \theta)
\end{equation}
This can be solved using \texttt{rootfinder} by defining $\mathbf{f}(\mathbf{y},\theta) = \mathbf{y} - \mathbf{g}(\mathbf{y},\theta)$.

\subsection{\texttt{minimize}}
Functional \texttt{minimize} solves the following equation,
\begin{equation}
    \mathbf{y^*} = \arg\min_\mathbf{y} f(\mathbf{y}, \theta)
\end{equation}
where $\mathbf{y}$ is a vector with size $n\times 1$ (practically it can be a tensor of any size), $\theta$ represents other argument, and $f$ is a function where the output is a scalar.

The vector-Jacobian product can be obtained from the rootfinder by substituting $\mathbf{f}\rightarrow \partial f/\partial \mathbf{y}$,
\begin{align}
    \mathbf{h}^T &= \frac{\partial \mathcal{L}}{\partial \mathbf{y}} \left(\frac{\partial^2 f}{\partial \mathbf{y}^2}\right)^{-1} \\
    \frac{\partial \mathcal{L}}{\partial \theta} &= -\mathbf{h}^T \frac{\partial^2 f}{\partial \mathbf{y} \partial \theta}.
\end{align}

\subsection{\texttt{quadrature}}

Functional \texttt{quadrature} performs 1-dimensional integration,
\begin{equation}
    \mathbf{y} = \int_{t_0}^{t_f} \mathbf{f}(t, \theta)\ \mathrm{d}t
\end{equation}
where $\mathbf{f}$ is a function with any size of outputs, $t$ is the integration variable, and $\theta$ represents other variables in $\mathbf{f}$.

The vector-Jacobian product is
\begin{align}
    \frac{\partial \mathcal{L}}{\partial \theta} &= \int_{t_0}^{t_1} \frac{\partial \mathcal{L}}{\partial \mathbf{y}} \frac{\partial \mathbf{f}}{\partial \theta}\ \mathrm{d}t.
\end{align}

\subsection{\texttt{solve\_ivp}}

Functional \texttt{solve\_ivp} solves the initial value problem (IVP) as expressed by
\begin{equation}
    \mathbf{y_1} = \mathbf{y_0} + \int_{t_0}^{t_1} \mathbf{f}(t,\mathbf{y}, \theta)\ \mathrm{d}t
\end{equation}
where $\mathbf{y_0}$ is a tensor of any size representing the initial condition, $\mathbf{f}$ is the function that calculates $\mathrm{d} \mathbf{y}/\mathrm{d}t$, $t_0$ and $t_1$ are respectively the initial and the final time, $\theta$ is other argument of $\mathbf{f}$, and $\mathbf{y_1}$ is the state at $t_1$.

The vector-Jacobian product are given by
\begin{align}
    \frac{\mathrm{d}\mathbf{a}(t)}{\mathrm{d}t} &= -\mathbf{a}(t)^T \frac{\partial \mathbf{f}(t, \mathbf{y}, \theta)}{\partial \mathbf{y}} \\
    \frac{\partial \mathcal{L}}{\partial \theta} &= \int_{t_1}^{t_0} \mathbf{a}(t)^T \frac{\partial \mathbf{f}(t,\mathbf{y},\theta)}{\partial \theta}\ \mathrm{d}t \\
    \frac{\partial \mathcal{L}}{\partial t_1} &=\mathbf{a}(t_1)^T\mathbf{f}(t_1,\mathbf{y_1},\theta) \\
    \frac{\partial \mathcal{L}}{\partial t_0} &=\frac{\partial \mathcal{L}}{\partial t_1} - \int_{t_1}^{t_0}\mathbf{a}(t)^T\frac{\partial \mathbf{f}(t,\mathbf{y},\theta)}{\partial t}\ \mathrm{d}t\\
    \frac{\partial \mathcal{L}}{\partial \mathbf{y_0}} &= \mathbf{a}(t_0)^T
\end{align}
where the first equation can be solved by integrating $\mathbf{a}(t)$ from $t_1$ down to $t_0$ with the following initial condition
\begin{equation}
    \frac{\partial \mathcal{L}}{\partial \mathbf{y_1}} = \mathbf{a}(t_1)^T
\end{equation}

\subsection{\texttt{mcquad}}

Functional \texttt{mcquad} integrates a function to get its expectation value over a probability distribution,
\begin{equation}
    \mathbf{y} = \mathbb{E}_p[\mathbf{f}(\mathbf{x}, \theta_f)] = \frac{\int \mathbf{f}(\mathbf{x}, \theta_f) p(\mathbf{x}, \theta_p)\ \mathrm{d}\mathbf{x}}{\int p(\mathbf{x}, \theta_p)\ \mathrm{d}\mathbf{x}}
\end{equation}
where $\mathbf{f}(\cdot)$ is the evaluation function, $p(\cdot)$ is the probability function which does not need to be normalized, $\theta_f$ and $\theta_p$ are the parameters for functions $\mathbf{f}(\cdot)$ and $p(\cdot)$ respectively

The vector-Jacobian product is
\begin{align}
    \frac{\partial \mathcal{L}}{\partial \theta_f} &= \mathbb{E}_p\left[\frac{\partial \mathcal{L}}{\partial \mathbf{y}}\frac{\partial \mathbf{f}}{\partial \theta_f}\right] \\
    \frac{\partial \mathcal{L}}{\partial \theta_p} &= \mathbb{E}_p\left[\frac{\partial \mathcal{L}}{\partial \mathbf{y}}(\mathbf{f} - \mathbf{y})\frac{\partial \log(p)}{\partial \theta_p}\right].
\end{align}

\end{document}